\documentclass{article}

\PassOptionsToPackage{numbers}{natbib}
 
\usepackage[final]{neurips_2025_ml4ps}

\usepackage[utf8]{inputenc} 
\usepackage[T1]{fontenc}    
\usepackage{hyperref}       
\usepackage{url}            
\usepackage{booktabs}       
\usepackage{nicefrac}       
\usepackage{microtype}      
\usepackage{xcolor}         

\usepackage{amsmath}
\usepackage{amssymb}
\usepackage{braket}
\usepackage{multirow}
\usepackage{graphicx}
\usepackage{dsfont}         
\usepackage{subcaption}

\graphicspath{ {./plots/} }

\title{Group Averaging for Physics Applications: \\ Accuracy Improvements at Zero Training Cost}

\author{%
  Valentino F.\ Foit \\
  Center for Cosmology and Particle Physics \\
  Department of Physics \\
  New York University \\
  New York, NY 10003 \\
  \texttt{foit@nyu.edu} \\
  \And
  David W.\ Hogg \\
  Center for Cosmology and Particle Physics \\
  Department of Physics \\
  New York University \\
  New York, NY 10003 \\
  \texttt{david.hogg@nyu.edu} \\
  \And
  Soledad Villar \\
  Department of Applied Mathematics \& Statistics \\
  and Mathematical Institute for Data Science \\
  Johns Hopkins University \\
  \texttt{svillar3@jhu.edu} \\
}

\begin{document}

\maketitle

\begin{abstract}
Many machine learning tasks in the natural sciences are precisely equivariant to particular symmetries.
Nonetheless, equivariant methods are often not employed, perhaps because training is perceived to be challenging, or the symmetry is expected to be learned, or equivariant implementations are seen as hard to build.
Group averaging is an available technique for these situations.
It happens at test time; it can make any trained model precisely equivariant at a (often small) cost proportional to the size of the group; it places no requirements on model structure or training.
It is known that, under mild conditions, the group-averaged model will have a provably better prediction accuracy than the original model.
Here we show that an inexpensive group averaging can improve accuracy in practice.
We take well-established benchmark machine learning models of differential equations in which certain symmetries ought to be obeyed.
At evaluation time, we average the models over a small group of symmetries.
Our experiments show that this procedure always decreases the average evaluation loss, with improvements of up to 37\% in terms of the VRMSE.
The averaging produces visually better predictions for continuous dynamics.
This short paper shows that, under certain common circumstances, there are no disadvantages to imposing exact symmetries; the ML4PS community should consider group averaging as a cheap and simple way to improve model accuracy.
\end{abstract}

\section{Introduction}

One of the key contributions of machine learning to scientific communities has been emulators and surrogate models.
These are data-driven approximations, typically deep neural networks, that learn the input-output mapping of a traditional numerical differential equation solver or physics-based simulation.
Once trained, a surrogate model can predict the system's evolution at a fraction of the computational cost.
Our work investigates how these surrogate models can be improved by enforcing some of the exact symmetries inherent in the underlying physical processes or equations.

It has been a long-standing and powerful idea in machine learning that models engineered to respect the symmetries of a problem should perform better. This concept has been formalized and given a rigorous theoretical foundation \cite{elesedy2021provably, petrache2023approximation, tahmasebi2023exact, huang2023approximately}; that work proves a generalization benefit for a class of models that are equivariant to a group action.
In other words, a model that respects the symmetry will (under conditions) have a strictly lower test error than a model that does not.

To demonstrate this idea, we use the data and machine learning models presented in ``The Well'' \cite{ohana2024well}, which is a collection of datasets containing numerical simulations of a wide variety of spatiotemporal physical systems.
The project also includes trained state-of-the-art surrogate machine learning models.
In this paper, we apply a simple and fast group averaging procedure to these trained models to produce new, exactly equivariant models \cite{yarotsky2022universal, murphy2019relational}. 
We show that this procedure improves the accuracy of models without any additional training.
This both demonstrates the value of equivariance and also delivers equivariance (after the fact) at almost no cost.
Here we use group averaging to make a very simple point, but in practice one should use the frame averaging technique, which implements symmetries by averaging over a small subset of group elements \cite{puny2022frame}.

\paragraph{Our contributions}

This paper is essentially a \emph{demonstration} of the principle of group averaging, which transforms a generic machine learning model into a precisely equivariant model.
Our demonstration shows that group averaging is practical and does improve model accuracy in a real physics context.
  
We emphasize that group averaging overcomes all of the standard objections to the use of equivariant approaches in machine learning:
It places no constraints on model architecture or structure, and it places no burden on training.
The discrete groups we use here --- the dihedral groups --- are small.
For continuous groups we approximated the average by sampling only a small number of elements.
So group averaging is very inexpensive.

\section{Methods} \label{methods}

\subsection{Surrogate models}

Surrogate models estimate the state variables of a spatiotemporal dynamical system, such as a parameterized family of PDEs with fixed parameters or the observations of a physical system.
The solution to a system of $s$ state variables over a spatial domain $\mathbb{D}$ and for all times $t \ge 0$ is represented as $\mathbf{u}: \mathbb{D} \times [0, \infty) \to \mathbb{R}^s$.

The domain $\mathbb{D}$ is typically a subset of two- or three-dimensional space, such as $[0, L_x] \times [0, L_y]$ or $[0, L_x] \times [0, L_y] \times [0, L_z]$.
We employ a standard uniform discretization of the system in time and space and denote by $N_D$ the number of elements in the discretized domain $\mathbb{D}$.
A snapshot $\mathbf{u}_t: \mathbb{R} \to \mathbb{R}^{s \times N_D}$ represents the values of all $s$ state variables at all $N_D$ points in the discretized domain at a fixed time $t$.

The surrogate model $M: \mathbb{R}^{k \times s \times N_D} \to \mathbb{R}^{s \times N_D}$ is an autoregressive prediction problem that maps a sequence of $k$ snapshots $\mathbf{U}_{t_i} = [ \mathbf{u}_{t_{i-k+1}}, \ldots, \mathbf{u}_{t_{i-1}}, \mathbf{u}_{t_i} ]$ to the snapshot of the following time step such that
\begin{align} \label{uhat}
    \mathbf{u}_{t_{i+1}} \approx \hat{\mathbf{u}}_{t_{i+1}} = M(\mathbf{U}_{t_i}).
\end{align}
The goal of the model $M$ is to minimize some distance between its predictions $\hat{\mathbf{u}}_t$ and the truth $\mathbf{u}_t$.
The initial conditions at $t_0$ are given by $k$ snapshots $\mathbf{U}_{t_0}$, so that the estimated next sequence is updated according to $\hat{\mathbf{U}}_{t_{1}} = [ \mathbf{u}_{t_{-k+2}}, \ldots, \mathbf{u}_{t_0}, \hat{\mathbf{u}}_{t_{1}} ]$.
The model $M$ is applied repeatedly to generate estimates for all desired times.

\subsection{Equivariance from group averaging} \label{group_averaging}

Our goal is to improve already-trained models that do not inherently respect some particular symmetry of the problem, by group averaging.
Let $G$ be a group. A function $f: X \to Y$ is $G$-equivariant if
\begin{align}
    f(\phi(g) x) = \psi(g) f(x) \quad \forall x \in X, \forall g \in G,
\end{align}
where $\phi$ is a representation of $G$ on $X$
and $\psi$ is a representation of $G$ on $Y$. In words, applying a transformation $\phi(g)$ to the input $x$ followed by the function $f$ is the same as applying the transformation $\psi(g)$ to $f(x)$. Equivariance is very common in physics, for example, all physically meaningful functions are equivariant with respect to coordinate transformations \cite{villar2023towards}.

Let $G$ be a compact group. The unique invariant measure $\lambda$ of $G$ is called the Haar measure. It is normalized $\lambda(G) = 1$, and invariant so that for every measurable subset $A \subset G$ and for any $g \in G$ it holds $\lambda(g A) = \lambda(A g) = \lambda(A)$.
The function $f$ can be projected onto the space of equivariant functions using the Reynolds operator defined as 
\begin{align} \label{proj}
    (\mathcal{Q} f)(x) = \int_G \psi(g^{-1}) f(\phi(g) x) ~ d \lambda(g).
\end{align}
This map essentially averages the function $f$ over the group transformations. In the case of finite groups, the integral is a sum over the group elements. In \cite{elesedy2021provably} it is proven that this averaging procedure reduces the error of the predictions. The assumptions to obtain provable improvements are: 
\begin{itemize}
    \item The underlying physical problem is $G$-invariant. This often holds because governing equations satisfy symmetries. 
    \item The distribution of the training data is $G$-invariant.
\end{itemize}

Our goal is to predict the dynamics of a system with a known symmetry. We can improve a surrogate model $M$ as in \eqref{uhat} with initial sequence $\mathbf{V}_{t_0} \equiv \mathbf{U}_{t_0}$ by group averaging
\begin{align} \label{vhat}
    \hat{\mathbf{v}}_{t_{i+1}} = \frac{1}{\lvert G \rvert} \sum_{g \in G} \psi(g^{-1}) M(\phi(g) \mathbf{V}_{t_i}).
\end{align}

Our approach approximates the continuous group average \eqref{proj} using a Monte Carlo method with $n$ random samples from its elements. We provide numerical examples for surrogate models in the next Section. We argue that this simple procedure provides an efficient, zero-shot method that allows us to inject physical knowledge into a model after it has been trained.

\section{Numerical experiments} \label{numexp}

We use the simulations and models published in ``The Well'' \cite{ohana2024well}. The project provides 15 TB of simulation data across 16 datasets, ranging from biological systems, fluid dynamics, acoustic scattering, as well as magneto-hydrodynamic simulations of extra-galactic fluids and supernova explosions.
We chose to analyze models corresponding to datasets defined on a Cartesian 2D grid with symmetric initial and boundary conditions.
For an overview of the groups used in this work, see Appendix \ref{groups}.
The dataset properties and state variables are summarized in Appendix \ref{expdetails}.

The \texttt{active\_matter} dataset \cite{maddu2024learning} consists of simulations of a continuum theory describing the dynamics of rod-like active particles immersed in a Stokes fluid.
The simulations were run at constant particle density and varying values for the parameters dimensionless active dipole strength and strength of particle alignment through steric interactions. The initial conditions of the quantities are plane-wave perturbations about the isotropic state.
The \texttt{gray\_scott\_reaction\_diffusion} dataset (from now on abbreviated as \texttt{gray\_scott}) contains numerical solutions of the Gray-Scott equations. They are a set of coupled reaction-diffusion equations that describe two chemical species, $A$ and $B$, whose concentrations vary in space and time. Two parameters control the ``feed'' and ``kill'' rates in the reaction. The initial conditions are either a random Fourier series or random clusters of Gaussians.
The boundary conditions are periodic for both datasets.
The underlying physical problems are sampled on a square grid and have the symmetry of the torus $\mathbb{T}^2$.
Toroidal symmetry allows all variables to translate periodically in both directions. Additionally, the problem is invariant under rotations and inversions, which is the discrete symmetry of the square $D_4$.

Additionally we analyze two datasets on a rectangular, non-square grid.
The boundary conditions are periodic in one direction and Neumann in the other.
The \texttt{rayleigh\_benard} dataset \cite{burns2020dedalus} contains simulations of convection between a hot and cold fluid layer. Initial random Gaussian noise creates density variations which causes fluid motion.
The underlying problem has the symmetry of the circle $S^1$ as well as inversions $D_2$.
The \texttt{turbulent\_radiative\_layer\_2D} dataset \cite{fielding2020multiphase} contains simulations of cold dense gas moving relative to hot dilute gas. Once turbulence is induced due to the Kelvin Helmholtz instability, mixing of the two layers occurs.
While the equations governing this process are invariant under translations and inversions, the initial conditions are not sampled from a distribution that is invariant under inversions. In all samples of the training set, the gas moves preferred in one direction. Consequently, the surrogate model does not respect this symmetry either.

The predictive power of the trained models vary greatly between the different datasets and models. Group averaging provides no improvement for the \texttt{shear\_flow} dataset because the specific initial conditions break the system's underlying symmetry, even though the theory itself and the boundary conditions are symmetric.

The CNextU-Net \cite{ronneberger2015u, liu2022convnet} was chosen in the analysis of the datasets due to its consistent performance.
The predictions $\hat{\mathbf{u}}_t$ and $\hat{\mathbf{v}}_t$ generated by \eqref{uhat} and \eqref{vhat} must be compared to the simulations $\mathbf{u}_t$. To measure how well a prediction fits the simulation, we employ the variance scaled root mean squared error (VRMSE)
\begin{align}
    \operatorname{\mathcal{L}_\text{VRMSE}}(\mathbf{u}, \mathbf{v}) = \left[
        \frac{\braket{\lvert \mathbf{u} - \mathbf{v} \rvert^2}}
        {\braket{\lvert \mathbf{v} - \braket{\mathbf{v}} \rvert^2} + \epsilon}
    \right]^{1/2},
\end{align}
where $\braket{\cdot}$ denotes the spatial mean operator, and $\epsilon = 10^{-7}$ to prevent division by zero in case $\operatorname{Var}(\mathbf{v}) = 0$. Lower $\operatorname{VRMSE}$ indicates higher agreement. $\operatorname{\mathcal{L}_\text{VRMSE}} \gtrsim 1$ signifies poor performance, as it indicates a prediction that is less accurate than a naively predicting the constant mean.

\subsection*{Results}

\begin{figure}[t]
    \small
    \centering
    \begin{subfigure}[b]{0.48\textwidth}
         \centering
         \includegraphics[width=\textwidth]{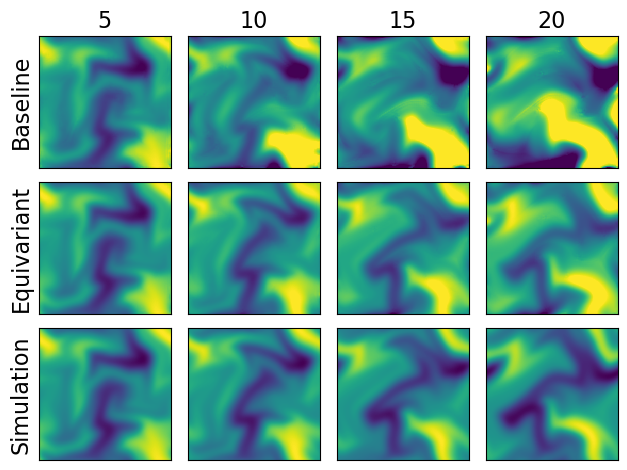}
         \caption{\texttt{active\_matter}}
         \label{active_matter_sim}
    \end{subfigure}
    \hfill
    \begin{subfigure}[b]{0.48\textwidth}
         \centering
         \includegraphics[width=\textwidth]{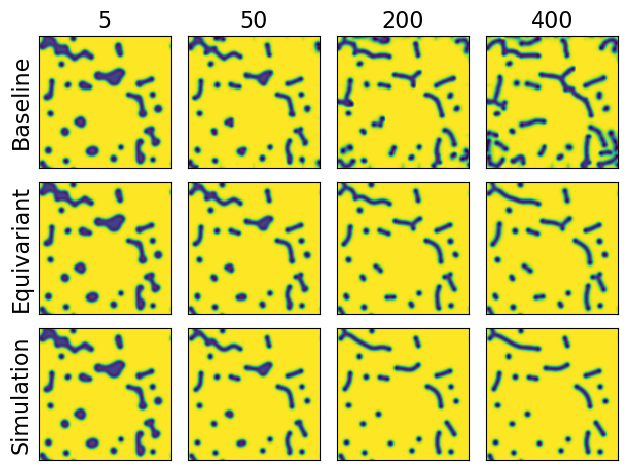}
         \caption{\texttt{gray\_scott}}
         \label{gray_scott_sim}
    \end{subfigure}
    \caption{\small Example comparisons of time evolutions of the concentration of two datasets.
    \subref{active_matter_sim} After $t=10$ steps we visually start to observe the benefits of group averaging. The equivariant model has a lower loss than the baseline at all times and is closer to the simulation (truth). At later times significant deviations between the approaches become apparent.
    \subref{gray_scott_sim} The equivariant model captures the features of the simulation relatively well for long times, while the baseline model develops several fictitious clusters. \label{simulations}}
    \vspace{12pt}
    \begin{subfigure}[t]{0.45\textwidth}
         \centering
         \includegraphics[width=\textwidth]{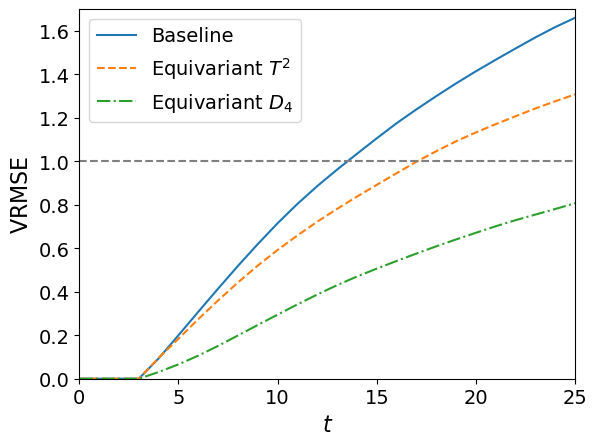}
         \caption{\texttt{active\_matter}}
         \label{active_matter_losses}
    \end{subfigure}
    \hfill
    \begin{subfigure}[t]{0.45\textwidth}
         \centering
         \includegraphics[width=\textwidth]{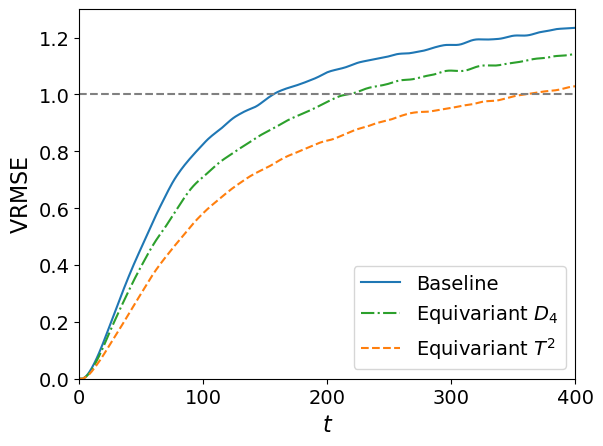}
         \caption{\texttt{gray\_scott}}
         \label{gray_scott_losses}
    \end{subfigure}
    \caption{\small Losses for the concentration predictions. In both cases the translations of $\mathbb{T}^2$ are performed with a single ($n=1$) random translation per time step. \label{losses}}
    \vspace{-10pt}
\end{figure}

We evaluate the performance improvement of the group averaging method by comparing the errors of the \textit{baseline} prediction
$\operatorname{\mathcal{L}_\text{VRMSE}}(\hat{\mathbf{u}}_t, \mathbf{u}_t)$
and of the \textit{equivariant} prediction
$\operatorname{\mathcal{L}_\text{VRMSE}}(\hat{\mathbf{v}}_t, \mathbf{u}_t)$.
Figure \ref{simulations} provides illustrations.
The results are summarized in Table \ref{table} and Figure \ref{losses}.

We experimented with the number of random Monte Carlo samples, $n$, used to approximate the continuous groups.
We observe significant improvements even at $n=1$, demonstrating an enhancement over the baseline surrogate models at no additional computational cost.
Table \ref{radiative_table} shows how the group average improves as $n$ increases.

The group averaging procedure outperforms the baseline prediction on all time frames, even if we sum over only one random representative of a continuous group per time step.
At sufficiently late times, the models lose predictive power and the loss approaches 1.
The rate of loss growth and the optimal averaging group are model-dependent.

\begin{table}[t]
    \scriptsize
    \renewcommand{\arraystretch}{1.1}
    \centering
    \caption{\small Average loss values for the datasets. The different starting positions in the trajectory and the used groups are given. The loss values are the averages of the $\operatorname{\mathcal{L}_\text{VRMSE}}$ over all trajectories and state variables in the respective test sets. The loss is evaluated at the time steps indicated above each column. The rollout loss is the sum of the first 15 errors after the initial sequence. The continuous symmetries $S^1$ and $\mathbb{T}^2$ are approximated with $n=1$ random group elements, except in \subref{radiative_table}, as indicated.}
    \label{table}
    \vspace{10pt}
    \centering
    \begin{subtable}[b]{0.48\textwidth} 
        \centering
        \begin{tabular}{@{}lccccc@{}}
            \toprule
            Model & Start & 1 & 5 & 10 & Rollout \\ \midrule
        
            Baseline & \multirow{3}{*}{10} & 0.063 & 0.386 & 0.938 & 10.81 \\
            $D_4$ & & \textbf{0.035} & \textbf{0.236} & \textbf{0.594} & \textbf{7.012} \\
            $\mathbb{T}^2$ & & 0.066 & 0.319 & 0.803 & 9.277 \\
            
            \midrule
            
            Baseline & \multirow{3}{*}{50} & 0.038 & 0.279 & 0.663 & 7.659 \\
            $D_4$ & & \textbf{0.026} & \textbf{0.193} & \textbf{0.497} & \textbf{5.821} \\
            $\mathbb{T}^2$ & & 0.038 & 0.202 & 0.507 & 5.976 \\
            
            \bottomrule
        \end{tabular}
        \caption{\texttt{active\_matter} \label{active_matter_table}}
    \end{subtable}~
    \begin{subtable}[b]{0.48\textwidth}
        \centering
        \begin{tabular}{@{}lccccc@{}}
            \toprule
            Model & Start & 1 & 10 & 50 & Rollout \\ \midrule
            
            Baseline & \multirow{3}{*}{10} & 0.008 & 0.154 & 0.742 & 1.833  \\
            $D_4$ & & \textbf{0.007} & 0.146 & 0.680 & 1.730 \\
            $\mathbb{T}^2$ & & 0.008 & \textbf{0.126} & \textbf{0.598} & \textbf{1.511} \\
            
            \midrule
            
            Baseline & \multirow{3}{*}{50} & 0.004 & 0.073 & 0.530 & 0.879 \\
            $D_4$ & & \textbf{0.003} & 0.064 & 0.453 & 0.764 \\
            $\mathbb{T}^2$ & & 0.004 & \textbf{0.051} & \textbf{0.357} & \textbf{0.608} \\
            
            \bottomrule
        \end{tabular}
        \caption{\texttt{gray\_scott} \label{gray_scott_table}}
    \end{subtable}
    \\
    \begin{subtable}[b]{0.48\textwidth}
        \centering
        \begin{tabular}{@{}lccccc@{}c}
            \toprule
            Model & Start & 1 & 10 & 20 & Rollout \\ \midrule
        
            Baseline & \multirow{3}{*}{50} & 0.065 & 0.346 & 0.627 & 4.246 \\
            $D_2$ & & \textbf{0.054} & 0.311 & 0.589 & 3.795 & \\
            $S^1$ & & 0.065 & \textbf{0.294} & \textbf{0.526} & \textbf{3.664} \\
            
            \midrule
            
            Baseline & \multirow{3}{*}{100} & 0.035 & 0.243 & 0.436 & 2.908 \\
            $D_2$ & & \textbf{0.032} & 0.216 & 0.394 & 2.586 \\
            $S^1$ & & 0.034 & \textbf{0.198} & \textbf{0.363} & \textbf{2.395} \\
            
            \bottomrule
        \end{tabular}
        \caption{\texttt{rayleigh\_benard}}
        \end{subtable}~
        \begin{subtable}[b]{0.48\textwidth}
        \centering
        \begin{tabular}{@{}lccccc@{}c}
            \toprule
            Model & Start & 1 & 5 & 10 & Rollout \\ \midrule
        
            Baseline & \multirow{3}{*}{0} & \textbf{0.142} & 0.350 & 0.579 & 7.834 \\
            $S^1_{n=1}$ & & 0.171 & 0.355 & 0.537 & 7.813 \\
            $S^1_{n=8}$ & & 0.144 & \textbf{0.306} & \textbf{0.487} & \textbf{6.911} \\
            
            \midrule
            
            Baseline & \multirow{3}{*}{50} & 0.201 & 0.499 & 0.744 & 9.128 \\
            $S^1_{n=1}$ & & 0.204 & 0.485 & 0.730 & 8.939 \\
            $S^1_{n=8}$ & & \textbf{0.174} & \textbf{0.449} & \textbf{0.722} & \textbf{8.605} \\
            
            \bottomrule
        \end{tabular}
        \caption{\texttt{turbulent\_radiative\_layer\_2D}}
        \label{radiative_table}
    \end{subtable}
\end{table}

\section{Discussion}

Imposing known symmetries theoretically improves the generalization error of machine learning models under mild conditions. However, in practice, equivariant models are not always better, maybe because it can be harder to train equivariant models than established non-equivariant models. 
Group averaging is a well-established alternative technique that delivers equivariance at no training cost and a post-processing cost that scales with the size of the group. 
Here we show that for certain physics applications it improves the models significantly.
Thus, when symmetry groups are small, there is no reason to sacrifice accuracy with non-equivariant surrogate models and emulators.

The dihedral groups we consider here are very small. The continuous groups we consider are approximated by sampling only a small number of elements. The methods used here could be improved for continuous groups, for example, by taking a subgroup or a frame of the continuous group \cite{puny2022frame}.
It remains to be seen if the average could be improved by using a Wasserstein barycenter, which is designed to preserve geometric structures in the data \cite{cuturi2014fast}.
Finally, many important groups for physics are not compact at all, such as translations, units transformations, boosts, coordinate diffeomorphisms, and variable reparameterizations.
Non-compact groups are out of scope for group averaging, though some averaging techniques can be extended to non-compact groups using the Weyl unitarian trick \cite{villar2023dimensionless}.

\newpage

\bibliographystyle{unsrt}
\bibliography{biblio}

\appendix

\section{Related work}

There has been a lot of recent work in machine learning for surrogate models, and they have demonstrated good performance in many domains \cite{stachenfeld2021learned,mccabe2023multiple, yang2023context, rahman2024pretraining, sun2025towards, shen2024ups, herde2024poseidon, chen2024omniarch}. These successes have motivated the creation of diverse datasets, such as \cite{takamoto2022pdebench, ohana2024well, mccabe2023multiple, audenaert2024multimodal}.

There is a vast literature on symmetry-preserving machine learning models \cite{shawe1989building,cohen2016group, kondor2018generalization}. Some authors have incorporated symmetry directly into the network architecture to create equivariant networks of different types \cite{worrall2019deep, wang2020incorporating, huang2023approximately, geiger2022e3nn, gregory2024robust}.
Several approaches to building equivariant models have been discussed, such as using invariant theory \cite{blum2023machine, villar2021scalars, blum2024galois, gregory2024learning}, group convolutions \cite{cohen2016group,cohen2019general}, canonicalization \cite{kaba2023equivariance}, and irreducible representations \cite{cohen2016steerable,kondor2018n}. In this work, we impose symmetries simply by averaging over the group orbit \cite{yarotsky2022universal, murphy2019relational}. This is a fairly naive approach that was significantly improved by recent work proposing \emph{frame averaging}, which only requires averaging over a subset of the group \cite{puny2022frame}.

Incorporating symmetries into machine learning models has been proven to improve generalization error and to reduce the sample complexity \cite{elesedy2021provably, petrache2023approximation, tahmasebi2023exact}. However, those are theoretical results that don't necessarily apply to complicated architectures such as those based on representation theory or group convolutions. In \cite{elesedy2021provably} it is mathematically shown that if the target function is equivariant, and the training data is sampled independently from an invariant distribution, then the projection \eqref{proj} reduces the generalization error.

\section{Groups} \label{groups}

We are interested in datasets on a square or rectangular Cartesian grid. Datasets defined on a square grid with appropriate initial and boundary conditions have the largest possible symmetry.

\subsection{Dihedral groups}

 The symmetry group of the square is $D_4$ and contains $\lvert D_4 \rvert = 8$ elements. For datasets defined on a rectangle, the biggest symmetry group is $D_2$, containing $\lvert D_2 \rvert = 4$ elements. Inversions correspond to $D_1 \cong C_2 \cong Z_2$ with $\lvert D_1 \rvert = 2$ elements.

Generally, the dihedral groups $D_n$ refer to the symmetry groups of the $n$-gon, groups of order $2n$. We can think of these as $n$ rotational symmetries and $n$ reflection symmetries.

The group representation of $D_4$ is
\begin{align}
    D_4 = \braket{r, i \vert r^4 = i^2 = (i r)^2 = e},
\end{align}
where its members can be thought of as rotations and inversions.
One representation is given by the matrices
\begin{align}
\begin{split} \label{matrep}
    R = \begin{pmatrix}
        0 & -1 \\
        1 & 0
    \end{pmatrix}, \quad
    I = \begin{pmatrix}
        -1 & 0 \\
        0 & 1
    \end{pmatrix},
\end{split}
\end{align}
and the $\lvert D_4 \rvert = 8$ group elements can be constructed by
\begin{align}
     \mathds{1}, R, R^2, R^3, I, I R, I R^2, I R^3.
\end{align}
$D_2$ can be constructed as above using only even powers of $R$, and $D_1$ by using only the inversion.

\subsection{Lie groups}

Translation invariant problems with periodic boundary conditions obey the symmetry of the circle.
The circle group is an example of a compact continuous Lie group. It is known by several names, such as $T \cong S^1 \cong U(1) \cong SO(2)$. The irreducible representations are one-dimensional maps to the unit circle.
If the problem is translation invariant and periodic in two directions, it is defined on a 2-torus. 
The torus group $\mathbb{T}^2$ is the continuous Abelian group formed from the direct product of two circle groups $\mathbb{T}^2 \cong S^1 \times S^1$.

Since numerical simulation data is always given on a discrete grid of dimension $N \times M$, we approximate the continuous groups with discrete cyclic groups $C_N \cong \mathbb{Z}/\mathbb{Z}_N$, that is $S^1$ with $C_N$ and $\mathbb{T}^2$ with $C_N \times C_M$.

\subsection{A vector is something that transforms like a vector}

We now look at how scalars, vectors, and tensors transform under the transformations of the aforementioned group elements. Since we are interested in datasets defined on a 2D Cartesian grid, 2-dimensional matrix representations of the group elements are useful.

Let us assume $\phi \equiv \phi(g)$ is a matrix representation of $g \in G$. 

Scalars do not transform
\begin{align}
    c \rightarrow c' = c.
\end{align}

Vectors transform like
\begin{align}
    \mathbf{u} \rightarrow \mathbf{u}' = \phi \mathbf{u}.
\end{align}

Tensors transform like
\begin{align}
    \mathbf{D} \rightarrow \mathbf{D}' = \phi \mathbf{D} \phi^\intercal.
\end{align}

For example, the orientation tensor $\mathbf{D}$ transforms under a $\pi/2$ rotation using \eqref{matrep} as
\begin{align}
\begin{split}
    \mathbf{D}' &= R \mathbf{D} R^\intercal =
    \begin{pmatrix}
        D_{yy} & - D_{yx} \\ - D_{xy} & D_{xx}
    \end{pmatrix}.
\end{split}
\end{align}
Unlike rotations or reflections, translations do not alter or mix the components of a vector or tensor. The entire fields are simply shifted.

\newpage

\section{Experimental details} \label{expdetails}

``The Well''\footnote{\url{https://polymathic-ai.org/the_well/}} \cite{ohana2024well} is published under \href{https://creativecommons.org/licenses/by/4.0/}{CC BY 4.0}. The baseline surrogate models were each trained on a specific dataset for a fixed duration of 12 hours on a single Nvidia H100 GPU.

Each test set contains $0.1 \times N$ of the trajectories. We excluded 9 trajectories from the \texttt{gray\_scott} dataset that converge to a steady state in our analysis.

\begin{table}[ht]
    \centering
    \renewcommand{\arraystretch}{1.1}
    \caption{Dataset specifications}
    \label{datasets}
    \vspace{10pt}
    \begin{subtable}[b]{\textwidth}
        \centering
        \begin{tabular}{@{}lccccc@{}}
            \toprule
            Dataset & Size (GB) & $N_D$ & $T$ & $N$ & $s$ \\ \midrule
            \texttt{active\_matter} & 51.3 & $256 \times 256$ & 81 & 360 & 11 \\
            \texttt{gray\_scott} & 153.8 & $128 \times 128$ & 1001 & 1200 & 2 \\
            \texttt{rayleigh\_benard} & 342 & $512 \times 128$ & 200 & 1750 & 4 \\
            \texttt{turbulent\_radiative\_layer\_2D} & 6.9 & $384 \times 128$ & 101 & 90 & 4 \\
            \bottomrule
        \end{tabular}
        \caption{Total size, resolution $N_D$, number of time steps per trajectory $T$, number of trajectories in the data set $N$, number of state variables $s$.}
        \label{data_specs}
    \end{subtable} \\
    \vspace{10pt}
    \begin{subtable}[b]{\textwidth}
        \centering
        \begin{tabular}{@{}llcc@{}}
            \toprule
            Dataset & Quantity & Type & Components \\ \midrule
            \multirow{4}{*}{\texttt{active\_matter}}
            & Concentration & Scalar & 1 \\
            & Velocity & Vector & 2 \\
            & Orientation & Tensor & 4 \\
            & Rate-of-strain & Tensor & 4 \\\midrule
            \multirow{2}{*}{\texttt{gray\_scott}}
            & Concentration $A$ & Scalar & 1 \\
            & Concentration $B$ & Scalar & 1 \\ \midrule
            \multirow{3}{*}{\texttt{rayleigh\_benard}}
            & Buoyancy & Scalar & 1 \\
            & Pressure & Scalar & 1 \\
            & Velocity & Vector & 2 \\ \midrule
            \multirow{3}{*}{\texttt{turbulent\_radiative\_layer\_2D}}
            & Buoyancy & Scalar & 1 \\
            & Pressure & Scalar & 1 \\
            & Velocity & Vector & 2 \\
            \bottomrule
        \end{tabular}
        \caption{The state variables of the datasets.}
        \label{statevariables}
    \end{subtable}
\end{table}

\end{document}